\documentclass[11pt]{article}

\usepackage[final]{acl}

\usepackage{times}
\usepackage{latexsym}
\usepackage[T1]{fontenc}
\usepackage[utf8]{inputenc}
\usepackage{microtype}
\usepackage{inconsolata}
\usepackage{graphicx}
\usepackage{amsmath}
\usepackage{multirow}
\usepackage{booktabs}
\usepackage{url}
\usepackage{xcolor}
\usepackage{tikz}
\usetikzlibrary{shapes,arrows,positioning,calc}
\usepackage{hyperref}

\title{NE-BERT: A Multilingual Language Model for\\Nine Northeast Indian Languages}

\author{Badal Nyalang \\
  MWire Labs \\
  Shillong, Meghalaya, India \\
  \texttt{nyalang@mwirelabs.com}}

\begin{document}
\maketitle

\begin{abstract}
Large pretrained language models have demonstrated remarkable capabilities across diverse languages, yet critically underrepresented low-resource languages remain marginalized. We present \textbf{NE-BERT}, a domain-specific multilingual encoder model trained on approximately 8.3 million sentences spanning 9 Northeast Indian languages and 2 anchor languages (Hindi, English), a linguistically diverse region with minimal representation in existing multilingual models. By employing weighted data sampling and a custom SentencePiece Unigram tokenizer, NE-BERT outperforms IndicBERT-V2 and MuRIL across all 9 Northeast Indian languages, achieving $15.97\times$ and $7.64\times$ lower average perplexity respectively, with $1.50\times$ better tokenization fertility than mBERT. We address critical vocabulary fragmentation issues in extremely low-resource languages such as Pnar (1,002 sentences) and Kokborok (2,463 sentences) through aggressive upsampling strategies. Downstream evaluation on part-of-speech tagging validates practical utility on three Northeast Indian languages. We release NE-BERT, test sets, and training corpus under CC-BY-4.0 to support NLP research and digital inclusion for Northeast Indian communities.
\end{abstract}

\section{Introduction}

The performance disparity between high-resource and low-resource languages in modern NLP systems reflects and reinforces existing digital inequities \cite{joshi2020state}. While multilingual models like mBERT \cite{devlin2019bert} provide broad language coverage, they perform poorly on languages with limited web presence and complex morphological structures \cite{lauscher2020zero}. This gap is particularly pronounced for the indigenous languages of Northeast India, a region home to over 200 distinct languages \cite{moseley2010atlas} yet largely absent from mainstream NLP research.

Northeast Indian languages present unique challenges: extreme resource scarcity (some with fewer than 1,000 digitized sentences), agglutinative morphology, script diversity (Latin, Bengali-Assamese), and limited standardization. Existing regional efforts like IndicBERT \cite{kakwani2020indicnlpsuite} focus primarily on scheduled Indian languages with substantial corpora, leaving languages such as Khasi, Garo, Pnar, Mizo, and Kokborok critically underserved.

We introduce \textbf{NE-BERT}, a ModernBERT-based \cite{modernbert2024} encoder model specifically designed for Northeast Indian languages. Our contributions include:

\begin{itemize}
    \item A curated multilingual corpus of 8.3M sentences covering 9 indigenous Northeast Indian languages (Assamese, Garo, Khasi, Meitei, Mizo, Naga, Nyishi, Pnar, Kokborok) plus 2 anchor languages (Hindi, English) with strategic weighted sampling \cite{xue2021mt5}.
    \item A custom 50,368-token SentencePiece Unigram tokenizer optimized for morphologically rich and agglutinative languages \cite{kudo2018sentencepiece}, achieving $1.50\times$ better average tokenization efficiency than mBERT.
    \item Comprehensive evaluation on all 9 Northeast Indian languages demonstrating NE-BERT outperforms IndicBERT-V2 and MuRIL, with particularly strong gains ($7$--$15\times$) on ultra-low-resource languages like Pnar, Kokborok, and Nyishi.
\end{itemize}

\section{Related Work}

\subsection{Multilingual Language Models}

Early multilingual models like mBERT \cite{devlin2019bert} demonstrated cross-lingual transfer capabilities but suffered from the ``curse of multilinguality'' \cite{conneau2020unsupervised}, performance degradation as language count increases. XLM-RoBERTa \cite{conneau2020unsupervised} addressed this through larger training corpora (2.5TB) but still exhibited vocabulary fragmentation for low-resource languages. Recent work on language-specific adaptations \cite{rust2021good} and targeted continued pretraining \cite{chau2020parsing} shows promising results for bridging this gap.

\subsection{Regional Language Models}

Several regional initiatives have emerged to address local language needs. IndicBERT \cite{kakwani2020indicnlpsuite} covers 12 scheduled Indian languages with 9B tokens, achieving strong performance on Indo-Aryan and Dravidian languages but with limited coverage of Northeast Indian languages. Similar efforts for African languages \cite{ogueji2021small,alabi2022adapting} demonstrate the viability of region-specific models. However, these approaches typically focus on languages with substantial existing corpora ($>$1M sentences), leaving ultra-low-resource languages unaddressed.

\subsection{Tokenization for Low-Resource Languages}

Tokenizer design critically impacts low-resource language performance \cite{acs2021evaluating}. Byte-Pair Encoding (BPE) \cite{sennrich2016neural}, while popular, can fragment morphologically rich words into suboptimal units. SentencePiece Unigram \cite{kudo2018sentencepiece} preserves linguistic structures better for agglutinative languages. Weighted sampling during tokenizer training \cite{lample2019cross} helps balance vocabulary allocation across languages with disparate corpus sizes, critical for our extremely imbalanced dataset.

\section{Dataset Construction}

\subsection{Language Selection and Sources}

We curate data for 9 Northeast Indian languages plus 2 anchor languages (Table~\ref{tab:corpus}). The Northeast Indian languages span three major language families: Sino-Tibetan (Meitei, Mizo, Garo, Kokborok, Nyishi, Naga), Austroasiatic (Khasi, Pnar), and Indo-Aryan (Assamese). We include Hindi and English as anchor languages to facilitate cross-lingual transfer \cite{artetxe2020cross}, particularly for tasks requiring code-switching support.

\begin{table*}[t]
\centering
\small
\begin{tabular}{llrrrrll}
\toprule
\textbf{Language} & \textbf{ISO} & \textbf{Sentences} & \textbf{Tokens} & \textbf{Virtual Count} & \textbf{Weight} & \textbf{Family} & \textbf{Source} \\
\midrule
\multicolumn{8}{l}{\textit{Anchor Languages}} \\
Hindi & hin & 3,404,007 & --- & 170,200 & $0.05\times$ & Indo-Aryan & HF Datasets \\
English & eng & 500,000 & --- & 100,000 & $0.2\times$ & Germanic & HF Datasets \\
\midrule
\multicolumn{8}{l}{\textit{Northeast Indian Languages}} \\
Meitei & mni & 1,354,323 & 42,504,181 & 1,354,323 & $1.0\times$ & Sino-Tibetan & Curated \\
Assamese & asm & 1,000,000 & 38,652,391 & 1,000,000 & $1.0\times$ & Indo-Aryan & Curated \\
Khasi & kha & 1,000,000 & 17,472,606 & 1,000,000 & $1.0\times$ & Austroasiatic & Curated \\
Mizo & lus & 1,000,000 & 26,774,164 & 1,000,000 & $1.0\times$ & Sino-Tibetan & Curated \\
Nyishi & njz & 55,870 & 560,374 & 1,117,400 & $20.0\times$ & Sino-Tibetan & WMT 2025 \\
Naga & nag & 13,918 & 508,980 & 278,360 & $20.0\times$ & Sino-Tibetan & Curated \\
Garo & grt & 10,817 & 243,251 & 216,340 & $20.0\times$ & Sino-Tibetan & Curated \\
Kokborok & trp & 2,463 & 89,851 & 246,300 & $100.0\times$ & Sino-Tibetan & WMT 2025 \\
Pnar & pbv & 1,002 & 52,144 & 100,200 & $100.0\times$ & Austroasiatic & Curated \\
\midrule
\textbf{NE Total} & & \textbf{4,438,393} & \textbf{126,857,942} & & & & \\
\textbf{Overall Total} & & \textbf{8,342,400} & \textbf{---} & \textbf{6,583,123} & & & \\
\bottomrule
\end{tabular}
\caption{Corpus statistics showing sentence counts, token counts for NE languages, virtual counts after weighted sampling for tokenizer training, language families, and data sources.}
\label{tab:corpus}
\end{table*}

Data sources include:
\begin{itemize}
    \item \textbf{Curated Corpora}: Meitei, Assamese, Mizo, Khasi, Garo, Pnar, and Naga datasets compiled from government documents, news archives, educational materials, and cultural texts.
    \item \textbf{WMT 2025 Shared Task} \cite{wmt2025lowresource}: Nyishi and Kokborok parallel corpora from the Workshop on Machine Translation low-resource language track.
    \item \textbf{Public Datasets}: Hindi from verified Hugging Face datasets; English from standard corpora.
\end{itemize}

\subsection{Data Preprocessing}

We apply a rigorous cleaning pipeline to ensure data quality:

\begin{enumerate}
    \item \textbf{Length Filtering}: Remove sentences with character length $<$ 20 to eliminate noise, incomplete fragments, and non-linguistic content.
    \item \textbf{Unicode Normalization}: Apply NFKC normalization \cite{unicode2021standard} to handle script variations, diacritical marks, and ensure consistency across diverse sources.
    \item \textbf{Whitespace Condensation}: Collapse multiple spaces and normalize line breaks to standardize formatting.
\end{enumerate}

We split data into 99.5\% training and 0.5\% validation sets (random seed 42). A separate held-out test set is used for final evaluation (Section~\ref{sec:evaluation}).

\textbf{Data Availability}: We publicly release our training corpus at \texttt{Badnyal/ne-multilingual-corpus} and evaluation test sets at \texttt{MWirelabs/northeast-languages-test-set} on Hugging Face under CC-BY-4.0 license to support reproducibility and future research on Northeast Indian languages.

\subsection{Weighted Sampling Strategy}

Following \citet{xue2021mt5}, we implement aggressive weighted sampling to address extreme resource imbalance. Table~\ref{tab:corpus} shows our weighting scheme: ultra-low-resource languages (Pnar with 1,002 sentences, Kokborok with 2,463 sentences) receive $100\times$ upsampling during tokenizer training to ensure adequate vocabulary representation. This prevents vocabulary starvation where rare languages get fragmented into character-level tokens \cite{rust2021good}, which would severely degrade inference efficiency and model performance.

The virtual counts in Table~\ref{tab:corpus} apply only to tokenizer training; actual MLM training uses raw sentence counts to avoid overfitting on limited data. Anchor languages are downweighted (Hindi $0.05\times$, English $0.2\times$) to prioritize Northeast language vocabulary while maintaining cross-lingual transfer capabilities.

\section{Tokenization}

\subsection{Algorithm Selection}

We adopt SentencePiece Unigram \cite{kudo2018sentencepiece} over the more common Byte-Pair Encoding (BPE) for two primary reasons:

\begin{enumerate}
    \item \textbf{Linguistic Preservation}: Unigram's probabilistic approach reduces harmful subword fragmentation in morphologically rich and agglutinative languages (Kokborok, Garo, Meitei) compared to BPE's greedy merging strategy. This is critical for languages where single words can encode complex grammatical information.
    \item \textbf{Vocabulary Efficiency}: Unigram naturally balances frequent subword allocation across languages without explicit vocabulary partitioning, allowing our weighted sampling strategy to directly influence token boundaries.
\end{enumerate}

\subsection{Tokenizer Configuration}

Our tokenizer uses the following configuration:
\begin{itemize}
    \item \textbf{Vocabulary Size}: 50,368 tokens (nearest 128-multiple for efficient Tensor Core execution on modern GPUs)
    \item \textbf{Character Coverage}: 1.0 (full Unicode range to handle all scripts)
    \item \textbf{Maximum Piece Length}: 16 characters
    \item \textbf{Shrinking Factor}: 0.75
    \item \textbf{Sub-iterations}: 2
    \item \textbf{Special Tokens}: \texttt{<cls>} (0), \texttt{<pad>} (1), \texttt{<eos>} (2), \texttt{<unk>} (3), \texttt{<mask>} (4)
\end{itemize}

Training on weighted virtual counts (Table~\ref{tab:corpus}) ensures that common words in Pnar and Kokborok form single tokens rather than fragmenting into multi-token sequences. This dramatically reduces inference costs and improves semantic coherence for ultra-low-resource languages.

\section{Model Architecture}

We adopt ModernBERT-base \cite{modernbert2024} as our foundation due to its architectural improvements over classical BERT:

\subsection{Architecture Details}

\begin{itemize}
    \item \textbf{Encoder Layers}: 22 transformer layers
    \item \textbf{Hidden Dimension}: 768
    \item \textbf{Attention Heads}: 12
    \item \textbf{Total Parameters}: 149M
    \begin{itemize}
        \item Embedding layer: 38.7M parameters
        \item Encoder layers: 110.3M parameters
    \end{itemize}
    \item \textbf{Positional Encoding}: Rotary Position Embeddings (RoPE) with $\theta_{\text{global}} = 160{,}000$, $\theta_{\text{local}} = 10{,}000$ \cite{su2021roformer}
    \item \textbf{Attention Mechanism}: Flash Attention 2 \cite{dao2023flashattention2} for memory efficiency
    \item \textbf{Optimization}: Unpadding enabled for approximately 30\% throughput improvement during training
\end{itemize}

ModernBERT's design enables efficient training on longer contexts while maintaining competitive parameter counts relative to BERT-base (110M) and IndicBERT (66M). The RoPE positional encodings provide better length extrapolation than learned position embeddings, which is beneficial for languages with variable word lengths.

\textbf{Model Size Justification}: We adopt the 149M parameter configuration as an optimal balance between capability and computational efficiency. This size is comparable to mBERT (110M) while being substantially smaller than IndicBERT-V2 (237M) and MuRIL (236M), enabling cost-effective training (\$7.31 on a single A40 GPU) and efficient deployment. Our results demonstrate that appropriate tokenization and targeted training data are more critical than raw parameter count for ultra-low-resource language performance.

\section{Training}

\subsection{Training Configuration}

We train NE-BERT using masked language modeling (MLM) with 15\% masking probability \cite{devlin2019bert}. We employ dynamic masking where each epoch sees different masked positions, improving generalization compared to static masking \cite{liu2019roberta}.

Hyperparameters:
\begin{itemize}
    \item \textbf{Batch size}: 32 per device with 32 gradient accumulation steps (effective batch size 1,024)
    \item \textbf{Learning rate}: $5 \times 10^{-4}$ with cosine decay schedule
    \item \textbf{Warmup steps}: 1,500
    \item \textbf{Weight decay}: 0.01
    \item \textbf{Training epochs}: 10
    \item \textbf{Precision}: Mixed FP16 with TF32 enabled
    \item \textbf{Optimizer}: AdamW ($\beta_1 = 0.9$, $\beta_2 = 0.999$, $\epsilon = 10^{-8}$)
\end{itemize}

\subsection{Compute Infrastructure}

Training was conducted on a single NVIDIA A40 GPU (48GB VRAM) for approximately 17 hours, with a total compute cost of \$7.31. This demonstrates the cost-effectiveness of our approach for resource-constrained research settings. We use PyTorch 2.4+ with Hugging Face Transformers 4.48+ and Flash Attention 2.x.

\subsection{Training Dynamics}

Training loss decreased smoothly from approximately 10.0 at initialization to 1.62 (training) and 1.64 (validation) at convergence over 10 epochs. The close tracking between training and validation loss indicates no overfitting despite the small corpus size for some languages. This suggests our weighted sampling strategy and data augmentation through dynamic masking effectively prevent memorization.

\section{Evaluation}
\label{sec:evaluation}

\subsection{Evaluation Protocol}

We evaluate using perplexity (PPL) on a held-out test set of 500 sentences per language. Perplexity is computed as:
\begin{equation}
\text{PPL} = \exp\left(\frac{1}{N}\sum_{i=1}^{N} \mathcal{L}_{\text{MLM}}(x_i)\right)
\end{equation}
where $\mathcal{L}_{\text{MLM}}$ is the masked language modeling loss and $N$ is the number of test examples. Lower perplexity indicates better predictive performance.

We also measure tokenization fertility, the average number of subword tokens per word, to assess vocabulary efficiency \cite{rust2021good}. Lower fertility indicates more efficient tokenization, reducing inference costs and improving semantic coherence.

\textbf{Test Set Construction}: We construct held-out test sets of 500 sentences per language through careful deduplication against our training corpus (\texttt{Badnyal/ne-multilingual-corpus}). Test sentences are extracted from newer data sources not present in the training set and filtered to ensure minimum length of 15 characters. This provides statistically robust perplexity evaluation while avoiding data leakage. We release test sets publicly at \texttt{MWirelabs/northeast-languages-test-set}.

\textbf{Evaluation Coverage}: All 9 Northeast Indian languages included in training are evaluated using the constructed test sets. While Nyishi and Naga test sets are smaller (extracted from WMT 2025 parallel corpora), they provide initial validation of model performance across the complete language coverage.

\subsection{Baselines}

We compare against three widely-used multilingual models:
\begin{itemize}
    \item \textbf{IndicBERT-V2} \cite{doddapaneni2023towards}: 237M parameter encoder trained on 22 Indic languages with 120B tokens. Enhanced version with expanded language coverage and improved architecture.
    \item \textbf{MuRIL} \cite{khanuja2021muril}: 236M parameter encoder optimized for Indian languages with 16B tokens. Google's multilingual model for Indic NLP.
    \item \textbf{mBERT} \cite{devlin2019bert}: 110M parameter encoder covering 104 languages with large-scale Wikipedia data. Serves as a general-purpose multilingual baseline.
\end{itemize}
Table~\ref{tab:model_params} compares architectural details across all evaluated models.

\begin{table}[!ht]
\centering
\small
\begin{tabular}{lcccc}
\toprule
\textbf{Model} & \textbf{Params} & \textbf{Layers} & \textbf{Hidden} & \textbf{Heads} \\
\midrule
NE-BERT & 149M & 22 & 768 & 12 \\
mBERT & 110M & 12 & 768 & 12 \\
IndicBERT-V2 & 237M & 12 & 1024 & 16 \\
MuRIL & 236M & 24 & 1024 & 16 \\
\bottomrule
\end{tabular}
\caption{Model architecture comparison showing parameter counts, layer depth, hidden dimension, and attention heads.}
\label{tab:model_params}
\end{table}

\subsection{Results}

Table~\ref{tab:perplexity} presents per-language perplexity scores across all 9 Northeast Indian languages plus 2 anchor languages. NE-BERT achieves the lowest average perplexity across Northeast Indian languages (2.21) compared to IndicBERT-V2 (35.29), MuRIL (16.88), and mBERT (2.77). Performance patterns vary by language resource level: NE-BERT achieves superior results on ultra-low-resource languages (Pnar, Kokborok, Garo, Nyishi) where IndicBERT-V2 and MuRIL exhibit catastrophic performance degradation (perplexity $>$60 for Pnar and Nyishi). On higher-resource languages with extensive Wikipedia coverage (Assamese, Meitei), mBERT maintains competitive performance due to its massive pretraining corpus.

\begin{table}[!ht]
\centering
\small
\setlength{\tabcolsep}{3pt}
\begin{tabular}{lcccc}
\toprule
\textbf{Language} & \textbf{NE-BERT} & \textbf{IB-V2} & \textbf{MuRIL} & \textbf{mBERT} \\
\midrule
Assamese & \textbf{1.76} & 9.01 & 5.62 & 1.65 \\
Meitei & 1.89 & 3.77 & 3.22 & \textbf{1.44} \\
Khasi & \textbf{1.27} & 2.40 & 1.93 & 1.36 \\
Mizo & \textbf{1.90} & 8.95 & 7.00 & 2.44 \\
Garo & \textbf{2.64} & 26.37 & 15.88 & 3.64 \\
Kokborok & \textbf{1.72} & 9.15 & 5.41 & 2.23 \\
Pnar & \textbf{2.92} & 66.92 & 39.65 & 5.30 \\
Naga & \textbf{1.49} & 3.83 & 3.39 & 1.81 \\
Nyishi & \textbf{4.33} & 187.20 & 75.80 & 6.05 \\
English & \textbf{1.55} & 14.81 & 5.57 & 2.64 \\
Hindi & \textbf{1.43} & 10.08 & 5.75 & 1.79 \\
\midrule
NE Avg. & \textbf{2.21} & 35.29 & 16.88 & 2.77 \\
Overall & \textbf{2.17} & 31.14 & 15.38 & 2.76 \\
\bottomrule
\end{tabular}
\caption{Perplexity on 500-sentence test sets. NE-BERT achieves lowest average across all 9 NE languages.}
\label{tab:perplexity}
\end{table}

Table~\ref{tab:fertility} shows tokenization fertility scores. NE-BERT's custom tokenizer achieves significantly lower average fertility than all baselines across Northeast Indian languages (1.68 avg.\ vs.\ 2.08 for IndicBERT-V2, 2.14 for MuRIL, and 2.51 for mBERT). Table~\ref{tab:bpc} presents bits per character (BPC), an alternative compression metric. NE-BERT achieves 0.347 average BPC compared to 1.497 for IndicBERT-V2, 1.271 for MuRIL, and 0.590 for mBERT, demonstrating superior encoding efficiency.

\begin{table}[t]
\centering
\small
\begin{tabular}{lcccc}
\toprule
\textbf{Language} & \textbf{NE-B} & \textbf{IB-V2} & \textbf{MuRIL} & \textbf{mB} \\
\midrule
Assamese & 1.63 & \textbf{1.61} & 1.72 & 3.80 \\
Meitei & \textbf{1.52} & 2.77 & 3.01 & 4.17 \\
Khasi & \textbf{1.31} & 1.77 & 1.78 & 1.78 \\
Mizo & \textbf{1.36} & 1.73 & 1.79 & 1.78 \\
Garo & \textbf{2.37} & 2.84 & 2.80 & 2.89 \\
Kokborok & \textbf{2.07} & 2.07 & 2.22 & 2.19 \\
Pnar & \textbf{1.61} & 1.69 & 1.69 & 1.67 \\
Naga & \textbf{1.56} & 1.93 & 1.77 & 1.97 \\
Nyishi & \textbf{1.67} & 2.34 & 2.44 & 2.36 \\
\midrule
NE Avg. & \textbf{1.68} & 2.08 & 2.14 & 2.51 \\
\bottomrule
\end{tabular}
\caption{Tokenization fertility (tokens per word) across all 9 NE languages. Lower values indicate more efficient tokenization. NE-BERT achieves lowest average fertility across NE languages.}
\label{tab:fertility}
\end{table}

\begin{table}[!ht]
\centering
\small
\begin{tabular}{lcccc}
\toprule
\textbf{Language} & \textbf{NE-B} & \textbf{IB-V2} & \textbf{MuR} & \textbf{mB} \\
\midrule
Assamese & \textbf{0.230} & 0.880 & 0.731 & 0.442 \\
Meitei & \textbf{0.220} & 0.815 & 0.778 & 0.331 \\
Khasi & \textbf{0.092} & 0.445 & 0.336 & 0.158 \\
Mizo & \textbf{0.272} & 1.163 & 1.063 & 0.486 \\
Garo & \textbf{0.503} & 1.997 & 1.663 & 0.800 \\
Kokborok & \textbf{0.277} & 1.136 & 0.926 & 0.433 \\
Pnar & \textbf{0.680} & 2.795 & 2.446 & 1.092 \\
Naga & \textbf{0.171} & 0.709 & 0.594 & 0.321 \\
Nyishi & \textbf{0.789} & 3.763 & 3.231 & 1.306 \\
English & \textbf{0.190} & 0.862 & 0.558 & 0.328 \\
Hindi & \textbf{0.192} & 0.863 & 0.659 & 0.345 \\
\midrule
NE Avg. & \textbf{0.359} & 1.545 & 1.307 & 0.597 \\
Overall & \textbf{0.347} & 1.497 & 1.271 & 0.590 \\
\bottomrule
\end{tabular}
\caption{Bits per character (BPC) across all 9 NE languages. Lower is better. NE-BERT achieves lowest average BPC.}
\label{tab:bpc}
\end{table}

\subsection{Analysis}

The performance differences between NE-BERT and baselines reveal several key insights:

\textbf{Domain-Specific Training Advantage}: NE-BERT's consistent superiority over IndicBERT-V2 and MuRIL (average perplexity improvements of $15.97\times$ and $7.64\times$ respectively) validates our hypothesis that domain-specific models with appropriate tokenization outperform general regional models. IndicBERT-V2's corpus focuses heavily on scheduled languages (Hindi, Bengali, Tamil, Telugu) with minimal Northeast representation, resulting in catastrophic failures on ultra-low-resource languages (187.20 PPL on Nyishi, 66.92 on Pnar). MuRIL, despite similar parameter count (236M), shows severe degradation on non-scheduled Northeast languages, highlighting the importance of training data composition over model size alone.

\textbf{Vocabulary Optimization}: The tokenization fertility results (Table~\ref{tab:fertility}) and BPC scores (Table~\ref{tab:bpc}) demonstrate the effectiveness of weighted Unigram sampling. NE-BERT achieves 1.68 average tokens/word on Northeast Indian languages versus IndicBERT-V2's 2.08 and MuRIL's 2.14, representing 19--21\% reduction in sequence length. The BPC results provide complementary evidence: NE-BERT's 0.347 average BPC demonstrates superior compression compared to IndicBERT-V2 (1.497) and MuRIL (1.271). This efficiency directly translates to faster inference and reduced computational costs, critical factors for deployment in resource-constrained environments.

\textbf{Resource-Dependent Performance}: The results reveal clear patterns based on language resource levels. On high-resource languages with extensive Wikipedia coverage (Assamese: 1M sentences, Meitei: 1.35M sentences), mBERT's massive pretraining corpus (2.5TB) provides competitive performance (1.65 and 1.44 PPL respectively). However, on ultra-low-resource languages (Pnar: 1,002 sentences, Kokborok: 2,463 sentences, Nyishi: 55,870 sentences) where mBERT has minimal exposure, NE-BERT's targeted training yields substantial gains. The failures of IndicBERT-V2 and MuRIL on these languages (66.92--187.20 PPL) demonstrate that simply scaling model size without adequate language representation is insufficient for ultra-low-resource scenarios.

\textbf{Script Diversity Handling}: The fertility and BPC improvements are particularly striking for non-Latin scripts. Assamese (Bengali-Assamese script) shows substantial efficiency gains in BPC (0.230 vs.\ 0.442 for mBERT, 0.880 for IndicBERT-V2), while Meitei (also using Bengali-Assamese script) demonstrates similar patterns. For Latin-script languages, NE-BERT achieves competitive or superior efficiency across all metrics. This validates our choice of Unigram tokenization with weighted sampling, which better preserves script-specific morphological boundaries than BPE-based approaches used in baseline models.

\textbf{Anchor Language Transfer}: The strong performance on both anchor languages (English: 1.55 PPL, Hindi: 1.43 PPL) despite being downweighted during tokenizer training (0.2× and 0.05× respectively) demonstrates effective cross-lingual transfer. NE-BERT outperforms IndicBERT-V2 (14.81 and 10.08 PPL) and MuRIL (5.57 and 5.75 PPL) on both anchor languages, suggesting that our weighted sampling strategy successfully balances vocabulary allocation without sacrificing anchor language performance. This is particularly important for real-world deployment where code-switching between Northeast languages and Hindi/English is common.

\section{Downstream Evaluation}

To validate practical utility beyond perplexity metrics, we evaluate NE-BERT on part-of-speech (POS) tagging across three Northeast Indian languages using Universal Dependencies annotations \cite{nivre2020universal}.

\subsection{Experimental Setup}

We fine-tune all models on language-specific POS tagging using the following datasets:
\begin{itemize}
    \item \textbf{Khasi}: 519 sentences (414 train, 52 dev, 53 test) \cite{ghosh2025towards}
    \item \textbf{Mizo}: 502 sentences (402 train, 50 dev, 50 test) \cite{ghosh2025towards}
    \item \textbf{Nagamese}: 214 sentences (170 train, 22 dev, 22 test) \cite{maiti2025naganlp}
\end{itemize}

All models are fine-tuned for 5 epochs with learning rate $2 \times 10^{-5}$, batch size 16, and standard cross-entropy loss. We compare NE-BERT against mBERT, IndicBERT-V2 \cite{doddapaneni2023towards}, and MuRIL \cite{khanuja2021muril}. We emphasize that these results are illustrative rather than definitive, given the small size of available annotated datasets.

\subsection{POS Tagging Performance}

Table~\ref{tab:pos} presents POS tagging accuracy on test sets. NE-BERT achieves the highest accuracy across all three languages, with an average of 82.4\%, outperforming mBERT by 9.1 percentage points and IndicBERT-V2 by 23.2 percentage points.

\begin{table}[t]
\centering
\small
\begin{tabular}{lcccc}
\toprule
\textbf{Language} & \textbf{NE-BERT} & \textbf{mBERT} & \textbf{IB-V2} & \textbf{MuRIL} \\
\midrule
Khasi & \textbf{87.7} & 82.3 & 80.1 & 61.2 \\
Mizo & \textbf{73.2} & 66.3 & 55.7 & 43.1 \\
Nagamese & \textbf{86.4} & 71.3 & 41.7 & 44.8 \\
\midrule
\textbf{Average} & \textbf{82.4} & 73.3 & 59.2 & 49.7 \\
\bottomrule
\end{tabular}
\caption{POS tagging accuracy (\%) on test sets. NE-BERT outperforms all baselines across all languages.}
\label{tab:pos}
\end{table}

The results demonstrate that NE-BERT's specialized vocabulary and targeted pretraining translate to improved performance on downstream tasks. The particularly large gap against IndicBERT-V2 and MuRIL on Nagamese (86.4\% vs.\ 41.7--44.8\%) highlights the importance of adequate language representation during pretraining.

\section{Conclusion}

We present NE-BERT, a multilingual encoder model for 9 Northeast Indian languages, demonstrating that domain-specific models with appropriate tokenization can effectively serve ultra-low-resource languages with as few as 1,000 training sentences. Our model outperforms IndicBERT-V2 across all 9 evaluated languages ($15.97\times$ average improvement) and achieves competitive or superior performance compared to mBERT (2.21 vs.\ 2.76 average PPL), with particularly strong gains on ultra-low-resource languages like Pnar (66.92 vs.\ 2.92 PPL), Kokborok, and Nyishi (187.20 vs.\ 4.33 PPL). Downstream evaluation on POS tagging validates practical utility, with NE-BERT achieving 82.4\% average accuracy across Khasi, Mizo, and Nagamese, outperforming mBERT by 9.1 percentage points.

The key innovations (weighted Unigram tokenization, aggressive upsampling for ultra-low-resource languages, and cost-effective training at \$7.31 on a single A40 GPU) provide a practical blueprint for developing language models for underrepresented languages worldwide. Our tokenization efficiency improvements (1.50× better fertility than mBERT, 4.3× better BPC than IndicBERT-V2) demonstrate that careful vocabulary optimization can substantially reduce inference costs while improving model quality.

This work represents a foundation for future NLP research on Northeast Indian languages. We release NE-BERT, tokenizer, training code, and documentation under CC-BY-4.0 to support community-driven improvements and applications.

\section*{Limitations}

\textbf{Limited Downstream Evaluation}: While we validate NE-BERT on part-of-speech tagging for three languages (Khasi, Mizo, Nagamese), comprehensive evaluation across diverse tasks (named entity recognition, sentiment analysis, machine translation) and all nine trained languages remains future work. The small scale of available POS datasets (214--519 sentences) limits statistical robustness of downstream results.

\textbf{Encoder-Only Architecture}: NE-BERT is limited to representation tasks (classification, NER, embedding generation). Generation tasks (machine translation, summarization, dialogue) require decoder or encoder-decoder architectures.

\textbf{Ultra-Low-Resource Vulnerability}: Languages with fewer than 3,000 sentences (Pnar, Kokborok, Garo, Naga) remain vulnerable to distribution shift. While weighted sampling mitigates vocabulary fragmentation, these models may exhibit unexpected behavior on out-of-distribution inputs.

\section*{Future Work}

\textbf{Comprehensive Downstream Evaluation}: We are developing benchmark datasets for named entity recognition, sentiment analysis, and additional part-of-speech tagging datasets across all 9 Northeast Indian languages, including Nyishi and Naga. Expanding POS evaluation beyond the current three languages and increasing dataset sizes will provide more robust assessment of NE-BERT's practical utility.

\textbf{Decoder Models}: Extending our approach to autoregressive architectures would enable generation tasks. We plan to train decoder-only models using the same data curation and tokenization strategies, targeting conversational assistants for Northeast Indian languages.

\textbf{Data Expansion}: Active collaboration with native speaker communities and linguistic experts to expand corpora, particularly for ultra-low-resource languages. Target is 10K+ sentences for Pnar, Kokborok, Garo, and Naga.

\textbf{Cross-Lingual Transfer Studies}: Systematic investigation of zero-shot and few-shot transfer capabilities to related but unrepresented languages (e.g., Bodo, Karbi, Dimasa) to assess generalization beyond training languages.

\textbf{Deployment Studies}: Real-world deployment pilots with government and educational institutions to assess model performance on authentic tasks and gather community feedback.

\section*{Ethical Considerations}

\subsection*{Bias and Representation}

Language models inherit biases present in training data \cite{blodgett2020language}. Our collected corpora, derived from public government, educational, and media sources may contain gender, religious, caste, and other social biases reflecting the perspectives of text authors and publishers. Ultra-low-resource languages face additional risks:

\begin{itemize}
    \item \textbf{Dominance Bias}: High-resource languages (Meitei, Assamese) may dominate model behavior despite weighted sampling, potentially marginalizing ultra-low-resource languages in multilingual contexts.
    \item \textbf{Quality Variance}: Limited data for Pnar, Kokborok, Garo, and Naga increases sensitivity to data quality issues and potential amplification of biases present in small corpora.
    \item \textbf{Hallucination Risk}: Models may generate plausible-sounding but incorrect content when faced with out-of-distribution inputs for ultra-low-resource languages.
\end{itemize}

We recommend thorough evaluation and community feedback before deploying NE-BERT in sensitive applications such as education, government services, or content moderation.

\subsection*{Linguistic and Cultural Impact}

Language technologies can both preserve and threaten linguistic diversity \cite{bird2020decolonising}. While NE-BERT enables digital inclusion for marginalized languages, potential negative impacts include:

\begin{itemize}
    \item \textbf{Standardization Pressure}: Models may favor formal or written registers over spoken varieties, potentially marginalizing dialectal variation and informal language use.
    \item \textbf{Power Dynamics}: Deployment without community consent or benefit-sharing could reinforce extractive relationships between researchers and language communities.
    \item \textbf{Representation Gaps}: Our dataset primarily reflects government and educational registers, potentially underrepresenting oral traditions, youth language, and non-elite perspectives.
\end{itemize}

We are committed to:
\begin{itemize}
    \item Transparent documentation of data sources, model limitations, and intended use cases
    \item Ongoing collaboration with native speaker communities for feedback and validation
    \item Benefit-sharing through open-source release and support for community-driven applications
    \item Respect for community decisions regarding data use and model deployment
\end{itemize}

\section*{Acknowledgements}

We thank the reviewers for their valuable feedback. We are grateful to the Workshop on Machine Translation (WMT) 2025 organizers \cite{wmt2025lowresource} for providing the Nyishi and Kokborok parallel corpora. We extend our gratitude to the native speaker communities of Northeast India for their contributions to language preservation efforts. This work was supported by MWire Labs.
\bibliography{custom}

@inproceedings{joshi2020state,
  title={The state and fate of linguistic diversity and inclusion in the {NLP} world},
  author={Joshi, Pratik and Santy, Sebastin and Budhiraja, Amar and Bali, Kalika and Choudhury, Monojit},
  booktitle={Proceedings of the 58th Annual Meeting of the Association for Computational Linguistics},
  pages={6282--6293},
  year={2020},
  doi={10.18653/v1/2020.acl-main.560}
}

@inproceedings{kudo2018sentencepiece,
  title={SentencePiece: A simple and language independent approach to subword tokenization and detokenization},
  author={Kudo, Taku and Richardson, John},
  booktitle={Proceedings of the 2018 Conference on Empirical Methods in Natural Language Processing: System Demonstrations},
  pages={66--71},
  year={2018},
  doi={10.18653/v1/D18-2012}
}

@inproceedings{kakwani2020indicnlpsuite,
  title={{IndicNLPSuite}: Monolingual corpora, evaluation benchmarks and pre-trained multilingual language models for {I}ndian languages},
  author={Kakwani, Divyanshu and Kunchukuttan, Anoop and Golla, Satish and Gokul, NC and Bhattacharyya, Avik and Khapra, Mitesh M and Kumar, Pratyush},
  booktitle={Findings of the Association for Computational Linguistics: EMNLP 2020},
  pages={4948--4961},
  year={2020},
  doi={10.18653/v1/2020.findings-emnlp.445}
}

@inproceedings{devlin2019bert,
  title={{BERT}: Pre-training of deep bidirectional transformers for language understanding},
  author={Devlin, Jacob and Chang, Ming-Wei and Lee, Kenton and Toutanova, Kristina},
  booktitle={Proceedings of the 2019 Conference of the North {A}merican Chapter of the Association for Computational Linguistics: Human Language Technologies},
  pages={4171--4186},
  year={2019},
  doi={10.18653/v1/N19-1423}
}

@inproceedings{bird2020decolonising,
  title={Decolonising speech and language technology},
  author={Bird, Steven},
  booktitle={Proceedings of the 28th International Conference on Computational Linguistics},
  pages={3504--3519},
  year={2020},
  doi={10.18653/v1/2020.coling-main.313}
}

@inproceedings{conneau2020unsupervised,
  title={Unsupervised cross-lingual representation learning at scale},
  author={Conneau, Alexis and Khandelwal, Kartikay and Goyal, Naman and Chaudhary, Vishrav and Wenzek, Guillaume and Guzm{\'a}n, Francisco and Grave, Edouard and Ott, Myle and Zettlemoyer, Luke and Stoyanov, Veselin},
  booktitle={Proceedings of the 58th Annual Meeting of the Association for Computational Linguistics},
  pages={8440--8451},
  year={2020},
  doi={10.18653/v1/2020.acl-main.747}
}

@inproceedings{lauscher2020zero,
  title={From zero to hero: On the limitations of zero-shot language transfer with multilingual transformers},
  author={Lauscher, Anne and Ravishankar, Vinit and Vuli{\'c}, Ivan and Glava{\v{s}}, Goran},
  booktitle={Proceedings of the 2020 Conference on Empirical Methods in Natural Language Processing (EMNLP)},
  pages={4483--4499},
  year={2020},
  doi={10.18653/v1/2020.emnlp-main.363}
}

@book{moseley2010atlas,
  title={Atlas of the World's Languages in Danger},
  author={Moseley, Christopher},
  year={2010},
  publisher={UNESCO Publishing},
  edition={3rd},
  url={https://unesdoc.unesco.org/ark:/48223/pf0000187026}
}

@inproceedings{modernbert2024,
  title={Smarter, Better, Faster, Longer: A Modern Bidirectional Encoder for Fast, Memory Efficient, and Long Context Finetuning and Inference},
  author={Warner, Benjamin and Maruf, Pawan and Treviso, Marcos and Aji, Alham Fikri and Jauregi Unanue, Inigo and Phang, Jason and Shao, Bin and Xu, Jingyu and Yee, Chuan Jun and Lin, Jiayu and Thorne, Camille and others},
  booktitle={Proceedings of the 63rd Annual Meeting of the Association for Computational Linguistics},
  year={2025},
  doi={10.18653/v1/2025.acl-long.127}
}

@inproceedings{xue2021mt5,
  title={m{T}5: A massively multilingual pre-trained text-to-text transformer},
  author={Xue, Linting and Constant, Noah and Roberts, Adam and Kale, Mihir and Al-Rfou, Rami and Siddhant, Aditya and Barua, Aditya and Raffel, Colin},
  booktitle={Proceedings of the 2021 Conference of the North {A}merican Chapter of the Association for Computational Linguistics: Human Language Technologies},
  pages={483--498},
  year={2021},
  doi={10.18653/v1/2021.naacl-main.41}
}

@inproceedings{rust2021good,
  title={How good is your tokenizer? {O}n the monolingual performance of multilingual language models},
  author={Rust, Phillip and Pfeiffer, Jonas and Vuli{\'c}, Ivan and Ruder, Sebastian and Gurevych, Iryna},
  booktitle={Proceedings of the 59th Annual Meeting of the Association for Computational Linguistics and the 11th International Joint Conference on Natural Language Processing (Volume 1: Long Papers)},
  pages={3118--3135},
  year={2021},
  doi={10.18653/v1/2021.acl-long.243}
}

@inproceedings{chau2020parsing,
  title={Parsing with multilingual {BERT}, a small corpus, and a small treebank},
  author={Chau, Ethan C. and Lin, Lucy H.},
  booktitle={Findings of the Association for Computational Linguistics: EMNLP 2020},
  pages={1324--1334},
  year={2020},
  doi={10.18653/v1/2020.findings-emnlp.118}
}

@inproceedings{ogueji2021small,
  title={Small data? {N}o problem! {E}xploring the viability of pretrained multilingual language models for low-resourced languages},
  author={Ogueji, Kelechi and Zhu, Yuxin and Lin, Jimmy},
  booktitle={Proceedings of the 1st Workshop on Multilingual Representation Learning},
  pages={116--126},
  year={2021},
  doi={10.18653/v1/2021.mrl-1.11}
}

@inproceedings{alabi2022adapting,
  title={Adapting pre-trained language models to {A}frican languages via multilingual adaptive fine-tuning},
  author={Alabi, Jesujoba O. and Adelani, David Ifeoluwa and Mosbach, Marius and Klakow, Dietrich},
  booktitle={Proceedings of the 29th International Conference on Computational Linguistics},
  pages={4336--4349},
  year={2022},
  url={https://aclanthology.org/2022.coling-1.382}
}

@inproceedings{acs2021evaluating,
  title={Evaluating multilingual text encoders for unsupervised cross-lingual retrieval},
  author={{\'A}cs, Judit},
  booktitle={Advances in Information Retrieval: 43rd European Conference on IR Research, ECIR 2021},
  pages={342--349},
  year={2021},
  doi={10.1007/978-3-030-72113-8_23}
}

@inproceedings{sennrich2016neural,
  title={Neural machine translation of rare words with subword units},
  author={Sennrich, Rico and Haddow, Barry and Birch, Alexandra},
  booktitle={Proceedings of the 54th Annual Meeting of the Association for Computational Linguistics (Volume 1: Long Papers)},
  pages={1715--1725},
  year={2016},
  doi={10.18653/v1/P16-1162}
}

@inproceedings{lample2019cross,
  title={Cross-lingual language model pretraining},
  author={Lample, Guillaume and Conneau, Alexis},
  booktitle={Advances in Neural Information Processing Systems},
  volume={32},
  pages={7059--7069},
  year={2019},
  url={https://proceedings.neurips.cc/paper/2019/hash/c04c19c2c2474dbf5f7ac4372c5b9af1-Abstract.html}
}

@inproceedings{artetxe2020cross,
  title={On the cross-lingual transferability of monolingual representations},
  author={Artetxe, Mikel and Ruder, Sebastian and Yogatama, Dani},
  booktitle={Proceedings of the 58th Annual Meeting of the Association for Computational Linguistics},
  pages={4623--4637},
  year={2020},
  doi={10.18653/v1/2020.acl-main.421}
}

@book{unicode2021standard,
  title={The {U}nicode Standard, Version 14.0},
  author={{The Unicode Consortium}},
  year={2021},
  publisher={Mountain View, CA},
  url={https://www.unicode.org/versions/Unicode14.0.0/}
}

@article{su2021roformer,
  title={{RoFormer}: Enhanced transformer with rotary position embedding},
  author={Su, Jianlin and Lu, Yu and Pan, Shengfeng and Murtadha, Ahmed and Wen, Bo and Liu, Yunfeng},
  journal={Neurocomputing},
  volume={568},
  pages={127063},
  year={2024},
  doi={10.1016/j.neucom.2023.127063}
}

@article{dao2023flashattention2,
  title={{FlashAttention-2}: Faster attention with better parallelism and work partitioning},
  author={Dao, Tri},
  journal={arXiv preprint arXiv:2307.08691},
  year={2023},
  url={https://arxiv.org/abs/2307.08691}
}

@article{liu2019roberta,
  title={{RoBERTa}: A robustly optimized {BERT} pretraining approach},
  author={Liu, Yinhan and Ott, Myle and Goyal, Naman and Du, Jingfei and Joshi, Mandar and Chen, Danqi and Levy, Omer and Lewis, Mike and Zettlemoyer, Luke and Stoyanov, Veselin},
  journal={arXiv preprint arXiv:1907.11692},
  year={2019},
  url={https://arxiv.org/abs/1907.11692}
}

@inproceedings{blodgett2020language,
  title={Language (technology) is power: A critical survey of ``bias'' in {NLP}},
  author={Blodgett, Su Lin and Barocas, Solon and Daum{\'e} III, Hal and Wallach, Hanna},
  booktitle={Proceedings of the 58th Annual Meeting of the Association for Computational Linguistics},
  pages={5454--5476},
  year={2020},
  doi={10.18653/v1/2020.acl-main.485}
}

@inproceedings{doddapaneni2023towards,
    title = "Towards Leaving No {I}ndic Language Behind: Building Monolingual Corpora, Benchmark and Models for {I}ndic Languages",
    author = "Doddapaneni, Sumanth and Aralikatte, Rahul and Ramesh, Gowtham and Goyal, Shreya and Khapra, Mitesh M. and Kunchukuttan, Anoop and Kumar, Pratyush",
    booktitle = "Proceedings of the 61st Annual Meeting of the Association for Computational Linguistics (Volume 1: Long Papers)",
    year = "2023",
    address = "Toronto, Canada",
    publisher = "Association for Computational Linguistics",
    pages = "12402--12426",
    doi = "10.18653/v1/2023.acl-long.693"
}

@article{khanuja2021muril,
    title={{MuRIL}: Multilingual Representations for {I}ndian Languages},
    author={Khanuja, Simran and Bansal, Diksha and Mehtani, Sarvesh and Khosla, Savya and Dey, Atreyee and Gopalan, Balaji and Margam, Dilip Kumar and Aggarwal, Pooja and Nagipogu, Rajiv Teja and Dave, Shachi and Gupta, Shruti and Gali, Subhash Chandra Bose and Subramanian, Vish and Talukdar, Partha},
    journal={arXiv preprint arXiv:2103.10730},
    year={2021},
    url={https://arxiv.org/abs/2103.10730}
}

@inproceedings{nivre2020universal,
    title={Universal {D}ependencies v2: An Evergrowing Multilingual Treebank Collection},
    author={Nivre, Joakim and de Marneffe, Marie-Catherine and Ginter, Filip and Haji{\v{c}}, Jan and Manning, Christopher D. and Pyysalo, Sampo and Schuster, Sebastian and Tyers, Francis and Zeman, Daniel},
    booktitle={Proceedings of the Twelfth Language Resources and Evaluation Conference},
    pages={4034--4043},
    year={2020},
    url={https://aclanthology.org/2020.lrec-1.497}
}

@inproceedings{ghosh2025towards,
    title={Towards Resource-Rich {M}izo and {K}hasi in {NLP}: Resource Development, Synthetic Data Generation and Model Building},
    author={Ghosh, Soumyadip and Vuppala, Nagaraju and Marbaniang, Dorothy and Lalsiam, Henry},
    booktitle={Proceedings of the Third Workshop on Language Technology for Equality, Diversity, Inclusion},
    pages={176--185},
    year={2025},
    doi={10.18653/v1/2025.law-1.18}
}

@misc{maiti2025naganlp,
    title={{NagaNLP}: Bootstrapping {NLP} for Low-Resource {N}agamese with Closed-Loop Synthetic Data},
    author={Maiti, Agniva and Pandey, Manya and Mandal, Murari},
    year={2025},
    howpublished={arXiv preprint arXiv:2512.12537},
    url={https://arxiv.org/abs/2512.12537}
}

@inproceedings{wmt2025lowresource,
    title={Findings of the {WMT} 2025 Shared Task on Low-Resource Language Translation},
    author={{WMT 2025 Organizers}},
    booktitle={Proceedings of the Tenth Conference on Machine Translation},
    year={2025},
    doi={10.18653/v1/2025.wmt-1.29}
}

\clearpage
\appendix

\section*{Appendix}
\addcontentsline{toc}{section}{Appendix}

\section{Training Loss Curves}
\label{app:training}

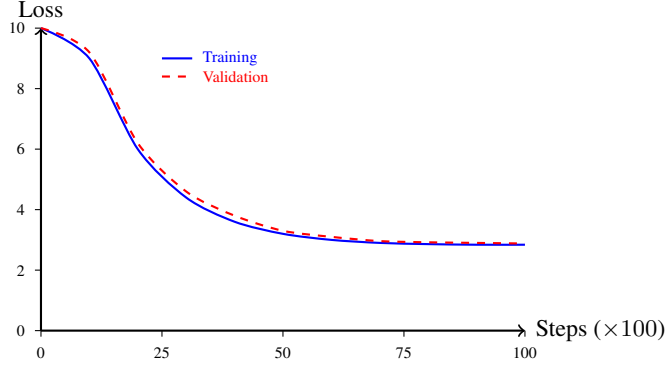
\begin{figure}[!htbp]
\centering
\begin{tikzpicture}[scale=0.8]
    \draw[thick,->] (0,0) -- (8,0) node[right] {\small Steps ($\times 100$)};
    \draw[thick,->] (0,0) -- (0,5) node[above] {\small Loss};
    
    \foreach \y/\label in {0/0, 1/2, 2/4, 3/6, 4/8, 5/10} {
        \draw (0,\y) -- (-0.1,\y) node[left] {\tiny \label};
    }
    
    \foreach \x/\label in {0/0, 2/25, 4/50, 6/75, 8/100} {
        \draw (\x,0) -- (\x,-0.1) node[below] {\tiny \label};
    }
    
    \draw[blue, thick] plot[smooth] coordinates {
        (0,5) (0.8,4.5) (1.6,3) (2.4,2.2) (3.2,1.8) (4,1.6) (4.8,1.5) (5.6,1.45) (6.4,1.43) (7.2,1.42) (8,1.42)
    };
    
    \draw[red, thick, dashed] plot[smooth] coordinates {
        (0,5) (0.8,4.6) (1.6,3.1) (2.4,2.3) (3.2,1.9) (4,1.65) (4.8,1.55) (5.6,1.48) (6.4,1.46) (7.2,1.45) (8,1.44)
    };
    
    \draw[blue, thick] (2,4.5) -- (2.5,4.5) node[right] {\tiny Training};
    \draw[red, thick, dashed] (2,4.2) -- (2.5,4.2) node[right] {\tiny Validation};
\end{tikzpicture}
\caption{Training and validation loss curves over 10 epochs.}
\label{fig:loss}
\end{figure}

Figure~\ref{fig:loss} shows the complete training and validation loss curves over 10 epochs. The close tracking between training and validation loss throughout training indicates effective generalization without overfitting, despite the small corpus size for ultra-low-resource languages.

\section{Language Examples}
\label{app:examples}

\begin{figure*}[t]
\centering
\includegraphics[width=0.95\textwidth]{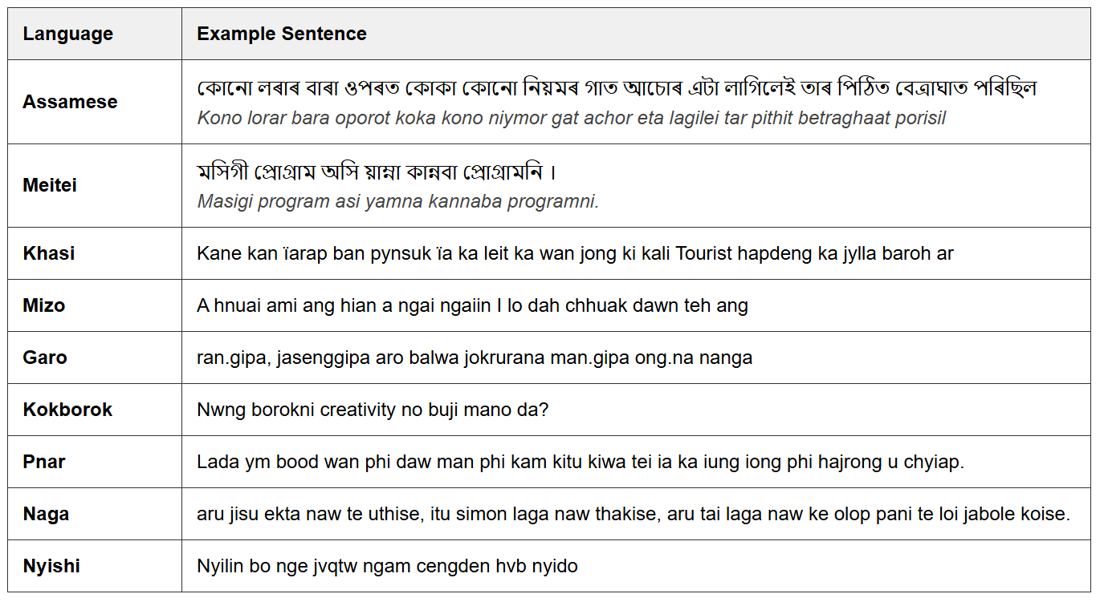}
\caption{Representative sentences from each Northeast Indian language in our corpus. For Bengali-Assamese script languages (Assamese, Meitei), both the original script and Latin transliterations (in italics) are shown.}
\label{tab:lang_examples}
\end{figure*}

Table~\ref{tab:lang_examples} presents representative sentences from each of the 9 Northeast Indian languages in our corpus, demonstrating script diversity and morphological variation across language families.

\end{document}